# A Comparative Study on Machine Learning Models to Classify Diseases Based on Patient Behaviour and Habits

**Elham Musaaed[1, 2, *], Nabil Hewah[2], Abdulla Alasaadi[3]**

[1] MSc. in Big Data Science and Analytics, College of Science, University of Bahrain, Al-Riffa, and Kingdom of Bahrain
Email: elham.musaad@gmail.com

[2] Department of Information Systems College of Information Technology, University of Bahrain, Al-Riffa, and Kingdom of Bahrain
Email: nhewahi@uob.edu.bh

[3] Department of Information Systems College of Information Technology, University of Bahrain, Al-Riffa, and Kingdom of Bahrain
Email: aalasaadi@uob.edu.bh

*Corresponding Author: Elham Musaaed, Email: elham.musaad@gmail.com

## Abstract

In recent years, ML algorithms have been shown to be useful for predicting diseases based on health data and posed a potential application area for these algorithms such as modeling of diseases. The majority of these applications employ supervised rather than unsupervised ML algorithms. In addition, each year, the amount of data in medical science grows rapidly. Moreover, these data include clinical and Patient-Related Factors (PRF), such as height, weight, age, other physical characteristics, blood sugar, lipids, insulin, etc., all of which will change continually over time. Analysis of historical data can help identify disease risk factors and their interactions, which is useful for disease diagnosis and prediction. This wealth of valuable information in these data will help doctors diagnose accurately and people can become more aware of the risk factors and key indicators to act proactively. The purpose of this study is to use six supervised ML approaches to fill this gap by conducting a comprehensive experiment to investigate the correlation between PRF and Diabetes, Stroke, Heart Disease (HD), and Kidney Disease (KD). Moreover, it will investigate the link between Diabetes, Stroke, and KD and PRF with HD. Further, the research aims to compare and evaluate various ML algorithms for classifying diseases based on the PRF. Additionally, it aims to compare and evaluate ML algorithms for classifying HD based on PRF as well as Diabetes, Stroke, Asthma, Skin Cancer, and KD as attributes. Lastly, HD predictions will be provided through a Web-based application on the most accurate classifier, which allows the users to input their values and predict the output. Logistic Regression (LR), Naive Bayes (NB), Random Forest (RF), K-Nearest Neighbor (KNN), Extreme Gradient Boost (XGB), and Support Vector Machine (SVM) were the algorithms used. The dataset was obtained from the Kaggle repository. The attributes are divided into PRF and diseases. The selected algorithms were implemented on the dataset with the optimal hyperparameters determined by using the "GridsearchCV" method in order to obtain the best performance. The accuracy of the algorithms ranged from 70% to 76%. Based on the accuracy, recall, precision, and F1-score measures for all algorithms, all ML algorithms predicted HD more accurately than diabetes, strokes, and KD. The algorithms even performed better when predefined diseases were combined with PRF in order to predict HD. Although there was no significant difference between the algorithms, LR achieved the highest score with 75%, when using only PRF and 76% when using a combination of disease attributes and PRF using a 70/30 split. Furthermore, accuracy increased from 74.8 to 76% when using the 10-fold CV. Two conclusions have been drawn: these features are more closely related to HD compared to other diseases and can be useful in predicting HD more proactively. Furthermore, the risk of HD increases with the presence of predefined diseases, especially Diabetes and Stroke. In terms of performance, LR was always one of the superior classifiers that performed similarly to more complex Machine Learning algorithms, while NB performed the worst.

**Keywords**

## 1. Introduction

There has been an increase in various diseases such as Diabetes and Heart Disease, as well as sudden deaths occurring at a younger age. Many factors contribute to the increase of various diseases, including People's lifestyles and environmental changes. According to the World Health Organization (WHO), there were 55.4 million deaths worldwide in 2019, with 55% attributed to the top 10 causes of death. Worldwide, three broad topics account for the majority of deaths: cardiovascular, respiratory, and neonatal conditions. As a general rule, death is classified according to three causes: communicable (infectious and parasitic diseases as well as maternal, perinatal, and nutritional conditions), non-communicable (chronic), and Injuries [1].

In WHO statistics, one of the biggest killers in the world is Ischemic Heart Disease (IHD), which is responsible for 16% of all deaths. In 2019, there have been 8.9 million deaths due to this disease, an increase of over 2 million since 2000. Moreover, about 11% and 6% of total deaths are caused by Stroke and Chronic Obstructive Pulmonary (COP) disease, respectively. Diabetes has risen 70% since 2000, becoming one of the top 10 causes of death. Male deaths due to Diabetes have increased by 80% since 2000, making it the leading cause of male death among the top 10. In addition, Kidney Disease is now the world's 10th leading cause of death, up from 13th. As of 2019, the death toll has increased to 1.3 million from 813,000 in 2000 [2].

Heart Disease, Stroke, and Cardiovascular risk factors are outlined in the American Heart Association's (AHA) annual reports, which include core health behaviors (smoking, physical activity, diet, and weight) and health factors (cholesterol, blood pressure, glucose control) related to cardiovascular health [1].

This calls for more awareness of the risks of our daily lifestyles and the impact they have on our health. Therefore, people need to be more aware of the risk factors and key indicators for such diseases due to the significant increase in death rates caused by these diseases, so they can act proactively and save lives [3]. Moreover, doctors may be unable to accurately predict a patient's condition based only on their symptoms. As there are several factors to consider when predicting diseases [4]. To enable medical professionals to make informed decisions regarding the occurrence of diseases, as well as to provide patients with a better understanding of how behavior and lifestyle impact their health, an accurate tool is needed.

Recently, supervised Machine Learning (ML) algorithms have shown to be useful for predicting diseases based on health data [5]. A considerable amount of research has been conducted to classify various diseases individually using ML algorithms. The majority of studies, however, have focused on clinical factors. Only a handful of studies have examined the effects of Diabetes in conjunction with clinical factors in predicting HD. Thus, this study provides insight into the relationship between a patient's behaviors and different diseases and how a patient's behavior and habits may contribute to a future medical condition. Furthermore, this study illustrated how Diabetes, Strokes, and KD - among other factors - contribute to HD. This study used supervised Machine Learning algorithms to classify big data and predict disease accurately based on behaviors and habits of patients that can be referred to as Patient-Related Factors. The Machine learning Algorithms are as follow Random Forest (RF), Support Vector Machine (SVM), Logistic Regression (LR), Naïve Bayes (NB), K-Nearest Neighbour (KNN), and Extreme Gradient Boost (XGB)).

The purpose of this study is to use ML approaches to fill this gap by conducting a comprehensive experiment to investigate the correlation between PRF and Diabetes, Stroke, HD, and KD. Moreover, it will investigate the relation between Diabetes, Stroke, KD, and PRF with HD. Further, the research aims to compare and evaluate various ML algorithms for classifying diseases based on the PRF. Additionally, it aims to compare and evaluate ML algorithms for classifying HD based on PRF associated with other diseases. Lastly, HD predictions will be provided through a Web-based application highlighting the most accurate classifier.

The remaining sections are arranged as follows: Literature Review, Exploratory Data Analysis (Data Collection, Data Summary & Data Processing), Methodology, Results, Discussion and Conclusion and Future Work.

## 2. Literature Review

Based on the conducted desktop research, it appears that the published literature addressing this topic is extremely rare and none of the papers cover the exact same issue that the vast majority of the papers are found to be focusing on predicting diseases based on clinical factors alone or by combining clinical factors with a few behavior habits. Only few studies have examined the effects of diabetes in conjunction with clinical factors in predicting HD. As a result, the authors of this paper sought to include literature that is closest to the research in hand.

A study published in 2020 examined Logistic Regression, Support Vector Machines, and Artificial Neural Networks (ANN) as algorithms for predicting Coronary Artery

Disease. A Z-Alizadeh Sani clinical dataset containing 303 patient records and 56 attributes was used to test the algorithms. Based on the test results, the ANN performed the best compared to the other algorithms with 93.35% accuracy [3].

In another study conducted by Mustaqeem et al. (2017), Cardiac Arrhythmia was classified based on 452 records and 279 features. As part of this study, a subset of features using a wrapper algorithm around the RF was selected, then implemented SVM, KNN, NB, RF, and Multi-Layer Perceptron (MLP) classifiers on these selected features. Results demonstrate that MLP surpasses KNN, SVM, and other algorithms by achieving 78.26% accuracy on average, while 76.6% and 74.4% accuracy are calculated for KNN and SVM, respectively [6].

Different ML algorithms were used by Dahiwade et al. (2019) to predict Heart Diseases. There were 303 records in the dataset with 75 features. In this study, the KNN algorithm and Convolutional Neural Network (CNN) were used. According to the results, CNN outperformed KNN by 87.5% [4].

A study by Uddin et al. (2019) identified those studies that used several supervised ML algorithms to predict a single disease. Various types of search items were searched in two databases (i.e., Scopus and PubMed). To compare variant supervised ML algorithms for disease prediction, 48 articles were selected. SVM was found to be more accurate in predicting three diseases (Heart Disease, Diabetes, and Parkinson's disease). In addition, SVM and NB were the most frequently used algorithms (in 29 and 23 studies, respectively). On the other hand, the RF algorithm demonstrated superior accuracy when compared to the other algorithms. In 53% of the studies utilizing RF, RF showed the highest accuracy. Moreover, SVM ranked first in 41% of the studies that used it [5].

In several studies, researchers used the publicly available Pima Indian Diabetes dataset from UCI for their experiments. There are 768 instances with 8 features in this dataset.

One of these studies was conducted by Patil et al. (2010). The authors propose the Hybrid Prediction Model (HPM) using the Simple K-means clustering algorithm in order to validate the classification label of the given data (incorrectly classified instances are removed) before applying the classification algorithm. Using the 10-fold cross-validation (CV) method, the final classifier model is constructed using the C4.5 algorithm. A classification accuracy of 92.38% was achieved with the proposed HPM [7].

In Nayak and Pandi (2021), different ML algorithms were applied to the same dataset. These included Decision Trees (DT), SVM, and KNN algorithms. They compared and analyzed the different accuracy measures. According to their experiment, the SVM algorithm achieved the highest accuracy at 73.95% [8].

According to Emon, et al. (2020), weighted voting classifiers have been developed to enhance the performance of Stroke prediction for doctors and patients such as detecting Stroke at an early stage. ML algorithms were compared to the proposed classifier and weighted voting provided the highest accuracy with 97% [9].

Rady and Anwar (2019) compared four Machine Learning algorithms to predict CKD, including Probabilistic Neural Networks (PNNs), MLP, SVM, and RBF. The aforementioned algorithms were tested using a dataset containing approximately 400 patient records with 25 attributes. According to the test results, the PNN performed best [10].

In another study, Ifraz et al. (2021) applied three different Machine Learning algorithms to the same dataset, selecting features based on a heat map, the absolute correlation between features, and the class label. A total of 14 attributes were used based on the feature selection. LR, DT, and KNN algorithms were used to analyze the data. The authors divided the data into training and testing segments, with training representing 80% of the data, and testing representing 20%. In their study, it was found that LR had a 97% accuracy rate over the other algorithms [11].

In most studies, clinical factors have been the primary focus. However, in this study, only Patient-Related Factors were considered, which include age, smoking, sleep time, general health, physical activity, physical health, mental health, alcohol consumption, difficulty walking, gender, race, and body mass index. Several diseases can be attributed to a person's lifestyle. Additionally, Diabetes, Stroke, and KD were examined in conjunction with PRF in order to assess their impact on the presence of HD.

## 3. Methods

### 3.1. Data Collection

The dataset was obtained from a secondary source. Specifically, it was downloaded from Kaggle Repository. It was maintained by the Disease Control and Prevention (CDC), which is a component of the Behavioral Risk Factor Surveillance System (BRFSS), that conducts telephone surveys annually to gather health-related data on residents of the United States. The datasets were in the Comma Separated Values (CSV) Format. Microsoft Excel was used to extract and view the dataset [12].

### 3.2. Data Summary

According to Table 1, the original dataset contains 319,000 records and 18 columns. The dataset contains 4 floats and 14 objects. A total of 17 attributes are considered to be featured, and the output will be a categorical attribute with a yes or no value indicating whether the individual has Heart Disease or not. Table 2 shows the distribution of these attributes. The attributes of a dataset can be divided into two categories: Patient-Related Factors and diseases. For the purpose of studying these factors, we have eliminated the diseases and kept the factors associated with each disease separately, resulting in four datasets: Heart, Diabetes, Stroke, and Kidney as illustrated in Table 3.

**Table 1** Original Dataset Attributes

| | | column | Datatype | Description |
|---|---|---|---|---|
| Output Features | patient-related factors | BMI | float | Body Mass Index |
| | | PhysicalHealth | float | How many days during the past 30 days was your physical health not good? (0-30 days) |
| | | SleepTime | float | Hours of sleeping in 24-hour period |
| | | MentalHealth | float | how many days during the past 30 days was mental health not good? (0-30 days) |
| | | GenHealth | Object | General health |
| | | Smoking | Object | Have you ever smoked? (Yes / No) |
| | | AlcoholDrinking | Object | Have you ever drank alcohol (Yes / No) |
| | | DiffWalking | Object | Difficulty walking or climbing stairs (Yes / No) |
| | | Sex | Object | Male or Female |
| | | AgeCategory | Object | Thirteen-level age category |
| | | Race | Object | Ethnicity |
| | | PhysicalActivity | Object | Doing physical activity or exercise during the past 30 days other than their regular job (Yes / No) |
| | Diseases | Asthma | Object | (Ever told) (you had) asthma? (Yes / No) |
| | | KidneyDisease | Object | (Ever told) (you had) kidney disease? (Yes / No) |
| | | SkinCancer | Object | (Ever told) (you had) skin cancer? (Yes / No) |
| | | Stroke | Object | (Ever told) (you had) a Stroke? (Yes / No) |
| | | Diabetic | Object | (Ever told) (you had) Diabetes? (Yes / No) |
| | | HeartDisease | Object | Have you ever had a heart attack? (Yes / No) |

**Table 2** Original Dataset Attributes Distribution

| Attribute | Distribution of records of Heart Disease | |
|---|---|---|
| Sex | Female = 41.1% | Male = 58.8% |
| Smoking | Yes = 58.6% | NO = 41.4% |
| AlcoholDrinking | Yes = 4.2% | NO = 95.8% |
| Stroke | Yes = 16.1% | NO = 83.9% |
| Diffwalking | Yes = 36.8% | NO = 63.2% |
| Asthma | Yes = 18.1% | NO = 81.9% |
| KidneyDisease | Yes = 12.7% | NO = 87.3% |
| SkinCancer | Yes = 18.2% | NO = 81.8% |
| AgeCategory | 18-24 = 0.7% | 25-29 = 0.8% |
| | 30-34 = 1.3% | 35-39 =1.5% |
| | 40-44 = 2.5% | 45-49 = 3.6% |
| | 50-54 = 5.9% | 55-59 = 8% |
| | 60-64 = 10.7% | 65-69 = 12.9% |
| | 70-74 = 16.5% | 75-79 =19.5% |
| | 80 or older = 23.27% | |
| Race | American Indian/Alaskan Native = 10.4% | |
| | Asian = 3.3% | Hipanic = 5.3% |
| | Black = 7.6% | Other = 8.14 |
| | White = 9.9% | |
| Diabetic | Yes = 33.2% | NO = 66.8% |
| GenHealth | Excellent = 2.5% | Very good = 5.1% |
| | Good = 10.5% | Fair = 20.5% |
| | Poor = 34.1% | |
| PhysicalActivity | Yes = 36.3% | NO = 63.8% |

**Table 3** Summary of Extracted Datasets

| Dataset | Features | Output (Predicting) |
|---|---|---|
| **Heart Dataset** | 12 Patient-Related factors | HeartDisease (Yes/No) |
| **Diabetic Dataset** | 12 Patient-Related factors | Diabetic (Yes/No) |
| **Stroke Dataset** | 12 Patient-Related factors | Stroke (Yes/No) |
| **Kidney Dataset** | 12 Patient-Related factors | Kidney (Yes/No) |

### 3.3. Data Processing

Data pre-processing is inevitable. The dataset must be carefully prepared prior to applying any ML algorithm. We followed these steps to analyze the data set: removing duplicated values, removing/replacing invalid or missing values, replacing outliers with the lower and upper limits - there were approximately 19000 records considered as outliers. We found that these records were valid, so we performed four scenarios and compared the results in order to determine which scenario was most accurate -, multiple-feature selection algorithms were implemented in order to select the most important features to be used in training the algorithms such as wrapper methods (forward/backward), RFECV methods, Balancing the dataset with the Random Under Sampler Method, standardizing the values with StandardScaler, and using One Hot Encoder for independent variables and Label Encoder for dependent variables. By using the encoder, the numerical data can be converted to categorical data. After processing the original data set, four datasets were extracted each consisting of 304,000 records with different output variables as shown in Figure 1. The same pre-processing steps were repeated for each extracted dataset to ensure that there are no outliers, balance the dataset, and that there are no duplicated records caused by dropping the unwanted attributes.

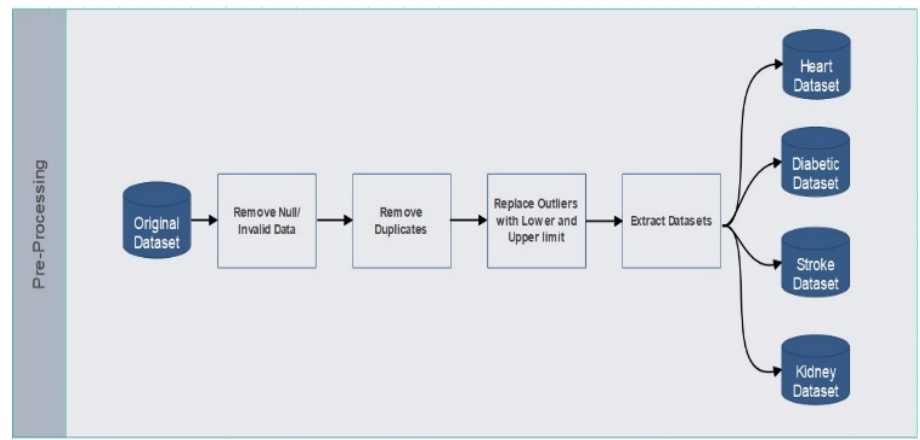

**Figure 1.** Pre-Processing Steps

### 3.4. Methodology

To conduct the research, multiple has been followed. Initially, exploratory data analysis and data pre-processing were conducted. After that, the original dataset was processed, and four datasets were extracted with different output variables. The Exploratory Data Analysis and data pre-processing were repeated for each extracted dataset. In order to select the most important features to be used in training the algorithms, multiple feature selection methods were implemented. Following the selection of the features and pre-processing of the dataset, two approaches were adopted. The first approach involved dividing the dataset into two parts, with 70% of the observations serving as a training set, while the remaining 30% as a test set. The second approach involved applying 10-fold Cross-Validation to the dataset. These datasets were used to train, validate, and test six different machine-learning

Elham Musaaed et al.

algorithms. Then the algorithms were evaluated in the final step. Figure 2 below illustrates the structure of the following methodology.

Using a pickle, the best algorithm is dumped. Next, a web application was developed using a flask application to enter input parameters. Hypertext Markup Language (HTML) code was used to develop the web page. In this application, the user enters input values to predict if Heart Disease will occur based on those input values. The parameters entered by the user are given to the flask application when the 'Submit' button is clicked.

Basically, a flask is a Python application that connects a web page with a trained ML algorithm. For prediction, the input values are sent to the flask application, which sends them to the algorithm, as shown in Figure 3.

We ran two different test data to test the effectiveness of the web page, one of which gave a prediction of 1 and the other gave a prediction of 0 as shown in Figure 4 and Figure 5.

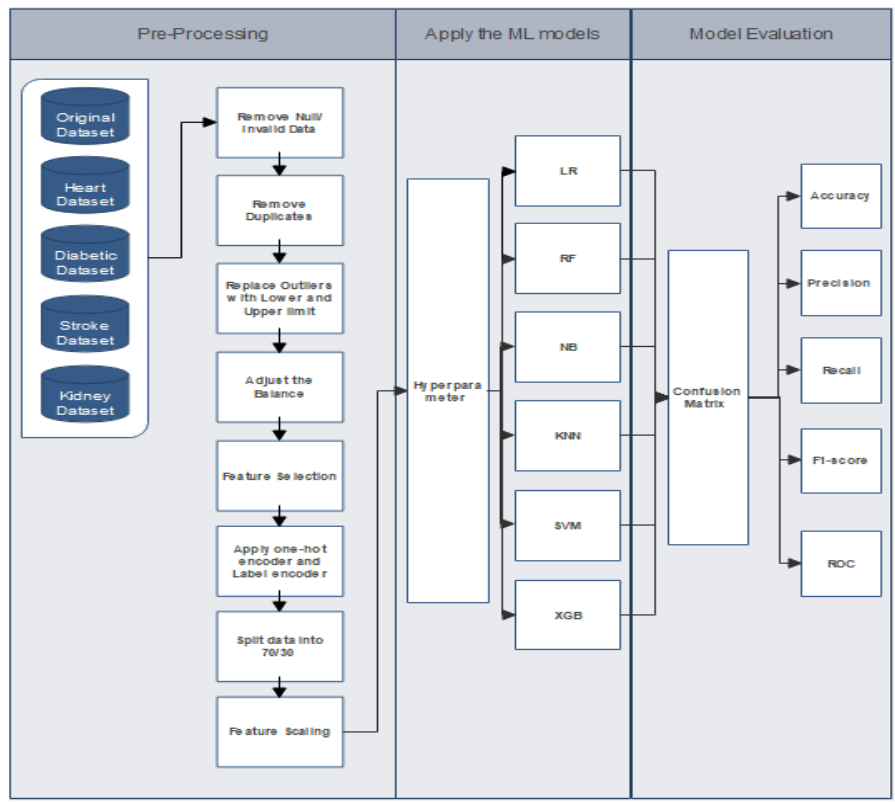

**Figure 2.** Research Methodology Structure Chart

**Figure 3.** Flask Application

**Figure 4.** Test Data predictions of 1

Output:
The User is more likely to have Heart Failuer

```
BMI                26.91
PhysicalHealth      0.00
MentalHealth        0.00
SleepTime           7.00
Smoking             0.00
AlcoholDrinking     0.00
DiffWalking         0.00
Sex                 1.00
AgeCategory         7.00
Race                5.00
PhysicalActivity    0.00
GenHealth           2.00
SkinCancer          0.00
Asthma              0.00
KidneyDisease       0.00
Stroke              0.00
Diabetic            1.00
Name: 16785, dtype: float64
```

Output:
The User is more likely to have healthy life

**Figure 5.** Test Data predictions of 0

### 3.4.1 Algorithm 1: RF

RF algorithm is capable of solving regression problems as well as classification problems. The algorithm works by choosing random samples with replacements to aggregate several base classifiers (Decision Trees). In classification, the prediction is made by a majority vote, while in regression; the prediction is made by averaging. As RF is a non-parametric algorithm, hence incorrect assumptions can be avoided. Additionally, it detects outliers and requires less data cleaning. The more we add trees, the higher the computational requirements become [12].

### 3.4.2 Algorithm 2: KNN

KNN is an algorithm that considers a non-parametric algorithm which solves both classification and regression problems. This algorithm is also referred to as lazy learning since the data is stored and the classification is not performed immediately. The algorithm involves selecting how many (k) neighbors are closest to a given point that will be considered in classifying it. Similarity is calculated in the KNN algorithm using the distance between two points. Accordingly, it will select neighbors based on the shortest distance between the given point and its neighbors. Points that are farther apart are less similar. The distance between two points can be calculated using a variety of methods, such as Euclidean, Manhattan, and Minkowski. However, Euclidean distance (i.e. the distance between two points in a straight line) is the most commonly used measurement [14]. This algorithm is highly influenced by the selected k value, which is the only hyperparameter. The determination of k is generally a matter of trial and error [15].

### 3.4.3 Algorithm 3: SVM

SVM is a supervised algorithm that is used to classify data. In order to classify data points, the algorithm creates a hyperplane in n-dimensional space to separate the classes of data points. The margin between classes is maximized using support vector points [16]. Among the advantages of SVM is its ability to handle non-linearly separable problems by applying kernel functions that transform the input space into a higher-dimensional space, thus converting the problem into a linearly separable one, consequently improving classification accuracy. Nevertheless, SVM can be slow when dealing with large datasets [17].

### 3.4.4 Algorithm 4: NB

NB algorithm is used to classify an outcome's instances. Using the Bayes theorem, it measures the probability of a dataset's values by calculating the frequency of their appearance [18]. Due to its high scalability, it has gained widespread acceptance. Furthermore, it can handle multi-class problems and missing values [19].

### 3.4.5 Algorithm 5: LR

LR is a classification algorithm that predicts a binary outcome based on a series of independent variables. It's a predictive classifying technique used to find the relationship between a dependent attribute and one or more independent attribute. It gives probability estimates that lie between 0 and 1. This classifier is known for its simplicity. It is an easy classifier to implement, train and interpret. The major drawback of this algorithm is that it assumes linearity between input and output [20].

### 3.4.6 Algorithm 6: XGB

XGB is an ensemble Supervised Learning algorithm that is based on gradient-boosted trees. A strong classifier is developed by combining predictions of weak classifiers in a serial training process. Learning can be accomplished more rapidly due to the parallel and distributed computations of the algorithm [21]. The process of sequentially growing trees using information from a previously developed tree is known as boosting. The algorithm gradually learns from the data and improves its forecasting abilities over time [22].

### 4. Results

The selected algorithms were first implemented on the extracted dataset with the optimal hyperparameters determined using the "GridsearchCV" method to obtain the best performance. In the second step, we applied them to the original dataset, which included Patient-Related Factors as well as Diabetes, Stroke, Asthma, Skin Cancer, and KD as attributes for predicting HD. In this section, we present the results from the experiments using 70/30 accuracy, 10-fold CV accuracy, and Receiver Operating Characteristic Area Under the Curve (ROC-AUC) for the original dataset as well as the extracted datasets. Therefore, we will be able to answer the following questions: How do Patient-Related Factors have an impact on Heart Disease, Diabetes, Strokes, and Kidney Disease? What are the effects of Diabetes, Stroke, and Kidney Disease associated with PRF on Heart Disease? and what is the most accurate algorithm?

In order for a classifier to be considered competent, the error rate (misclassified instances) must be low while recall (also known as sensitivity), specificity, and accuracy must be high. To further evaluate the performance of the chosen ML

Elham Musaaed et al.

algorithms, recall, precision, and F1-score were considered. Specifically, prediction precision indicates the quality of the algorithm in making accurate positive predictions, whereas prediction recall indicates the proportion of true positives. The increase in precision will result in a decrease in recall, and vice versa, so both cannot be maximized simultaneously in practice. In order to determine an algorithm's F1-score, precision, and recall are combined, two previously opposing metrics.

In Figure 6, we can see the results of applying LR. When using PRF to predict Heart Disease, the LR yields the best accuracy when compared with using PRF to predict Diabetes, Strokes, or kidneys, with 75% accuracy using 70/30 and 74.8% accuracy using 10-fold CV. With an accuracy of 73% using 70/30 and 73.1% using 10-fold CV, Stroke prediction comes second. Additionally, for Diabetics, the accuracy was 72% using 70/30 and 71.4% using 10-fold CV, while for Kidneys, the accuracy was 72% using 70/30 and 71.8% using 10-fold CV.

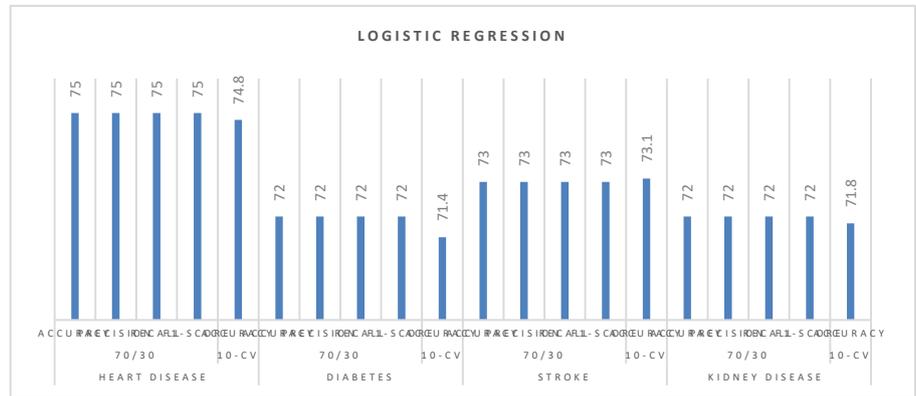

**Figure 6.** LR Results of Extracted Datasets Using Only Patient-Related Factors

As shown in Figure 7, when the original dataset is used, the accuracy of the LR increased from 74.8% to 76% with a 10-fold CV and from 75% to 76 % using a 70/30 split.

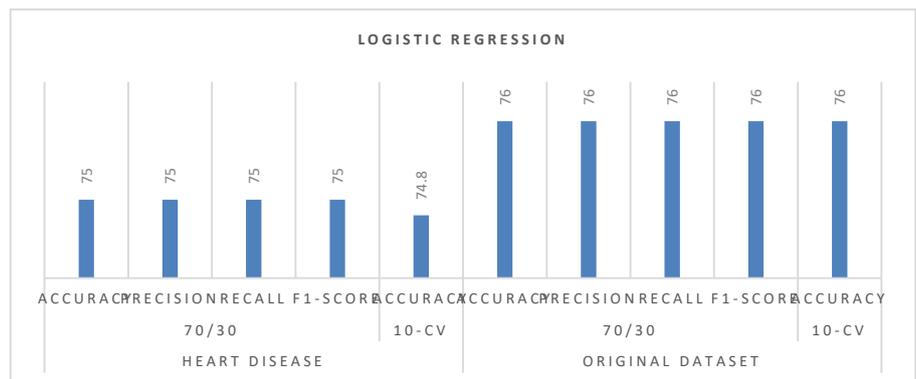

**Figure 7.** LR Results of Using PRFs with Disease to Predict HD

DOI: 10.33969/AIS.2024060103    46    Journal of Artificial Intelligence and Systems

In the results of NB in Figure 8, it can be noticed that the same behavior was even though there was a slight variation in the performance, however, Heart Disease was still the most accurate prediction with 72% accuracy using 70/30, while 71.8% accuracy when using a 10-fold CV. Further, the diabetic prediction was the lowest with 69% accuracy using 70/30, and 68.6% using 10-fold CV. When NB was applied to the original dataset, accuracy increased from 72% to 74% using 70/30, and accuracy was increased from 71.8% to 74% using a 10-fold CV as shown in Figure 9.

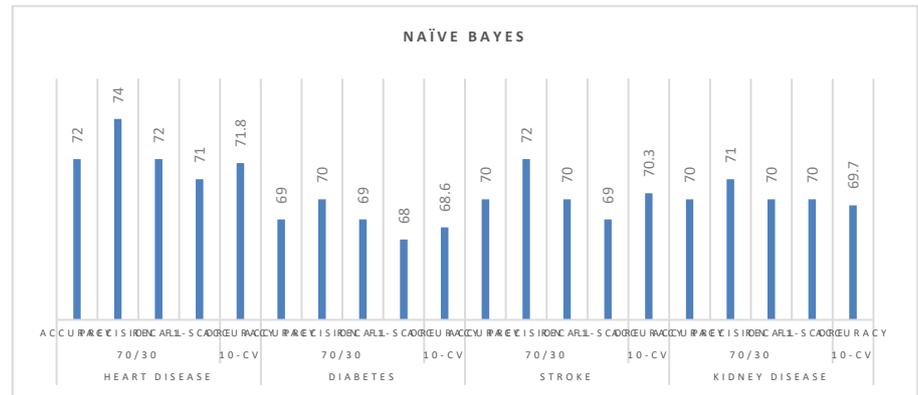

**Figure 8.** NB Results of Extracted Datasets Using Patient-Related Factors

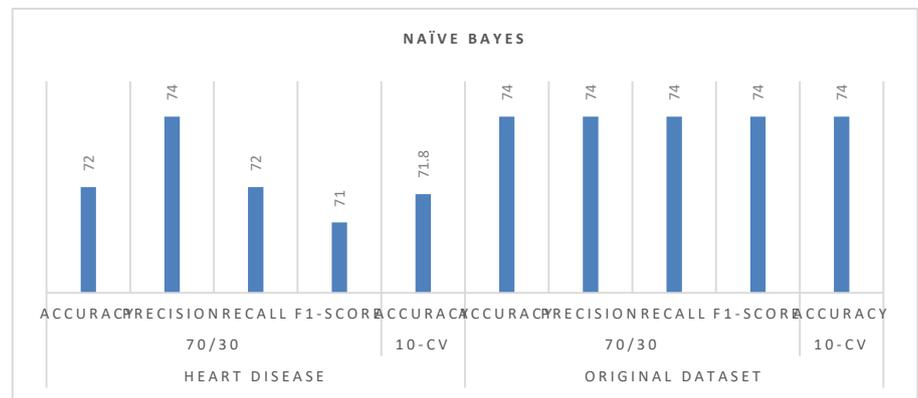

**Figure 9.** NB Results of the Original Dataset

In addition, the KNN algorithm provides the best performance when predicting HD with an accuracy of 74% using 70/30, and 74.6% using 10-fold CV. However, the accuracy for predicting Diabetes, Strokes, and KD were lower as shown in Figure 10. Moreover, Figure 11 illustrates the KNN performance when Diabetes, Stroke, and Kidney were included as attributes along with PRF. Based on 70/30 method, the performance - in terms of accuracy - slightly increased to 75%, while it increased from 74.6% to 75.2 % via the 10-fold CV method.

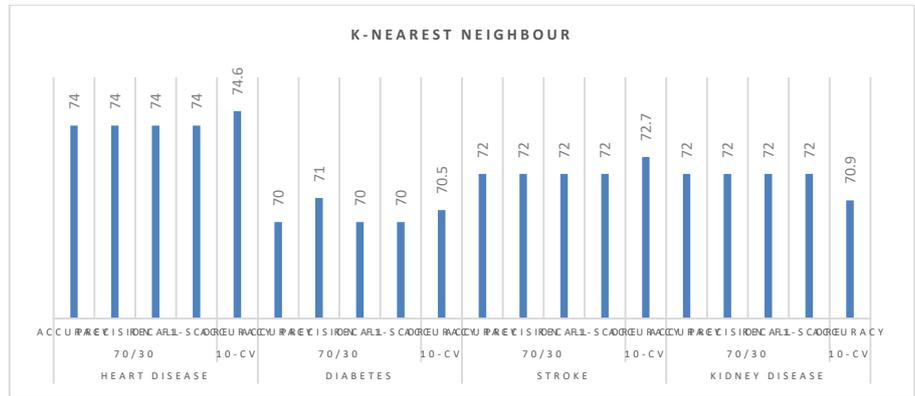

**Figure 10.** KNN Results of Extracted Datasets Using Patient-Related Factors

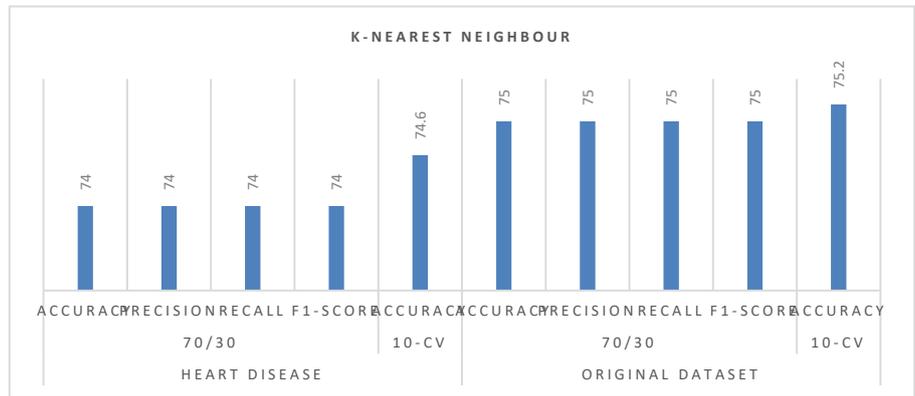

**Figure 11.** KNN Results of the Original Dataset

In Figure 12, the RF algorithm performed the best when used to predict HD with an accuracy of 75% using 70/30 and 74.8% using 10-fold CV. However, in other datasets, the percentage ranges between 71% and 73%. Nevertheless, when RF was applied to the original dataset as shown in Figure 13, 10-fold CV accuracy increased to 75.5% as opposed to using only PRF to predict HD.

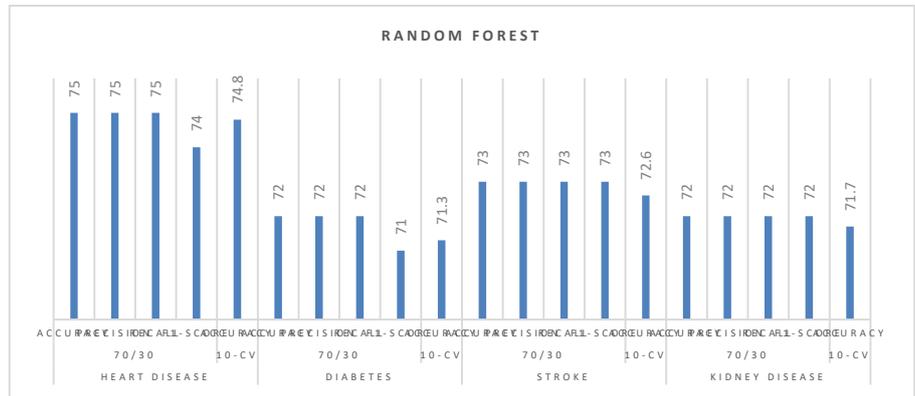

**Figure 12.** RF Results of Extracted Datasets Using Patient-Related Factors

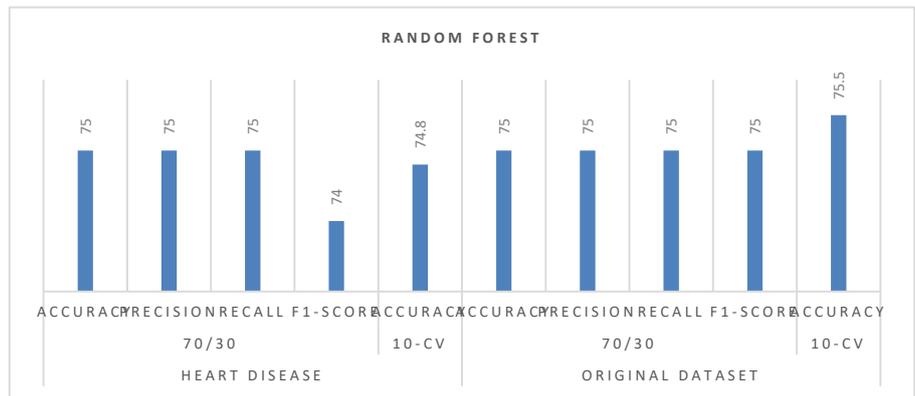

**Figure 13.** RF Results of the Original Dataset

Moving to the XGB results in Figure 14. It can be observed that when using PRF it performs best to predict HD with 74% accuracy using 70/30, and it gives 74.8% accuracy using a 10-fold CV. On the other hand, performance varies between 71.6% and 73 % for other extracted datasets. While in Figure 15, the 10-fold CV accuracy increased from 74.8% to 75.7% when applied to the original datasets.

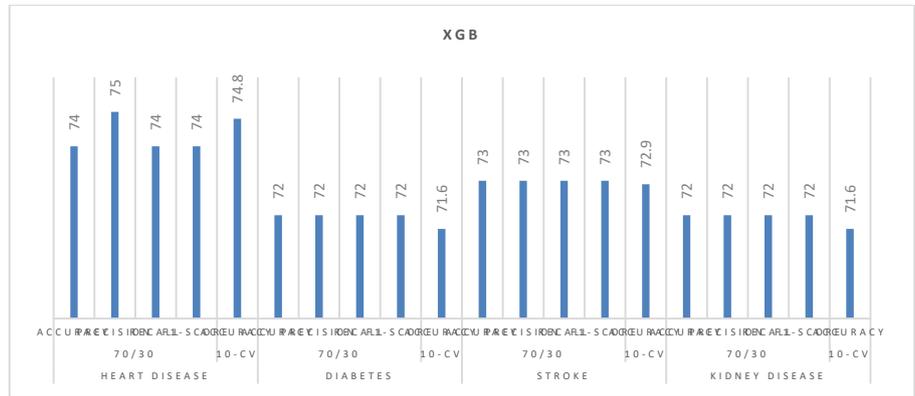

**Figure 14.** XGB Results of Extracted Datasets Using Patient-Related Factors

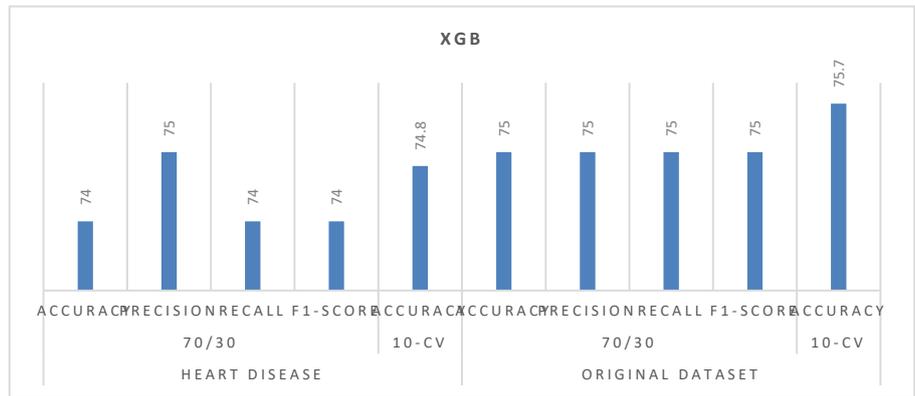

**Figure 15.** XGB Results of the Original Dataset

In comparison with the other algorithms used, the SVM gives the same result. According to Figure 16, it gives the highest accuracy when used to predict HD based on PRF, with 74% using 70/30 and 74.8% 10-fold CV. While the rest of the dataset showed an accuracy range of 70.4% to 73%. However, when it was used with the original dataset to predict HD, the accuracy improved for both 70/30 at 75% and 10-fold CV at 75.7% as shown in Figure 17.

Elham Musaaed et al.

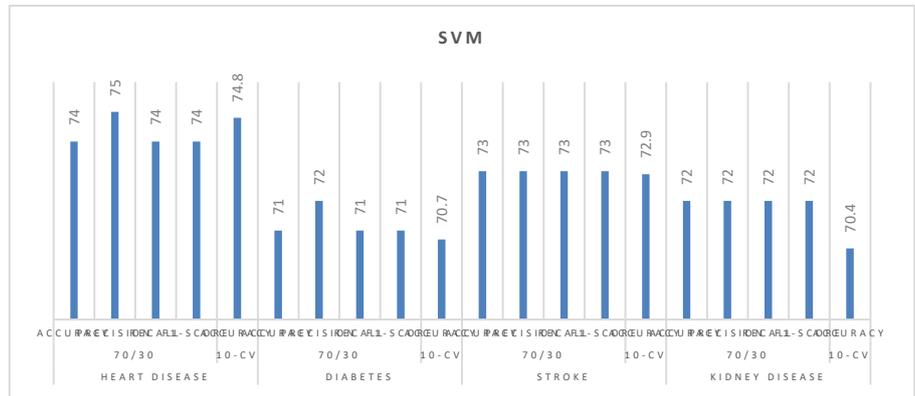

**Figure 16.** SVM Results of Extracted Datasets Using Patient-Related Factors

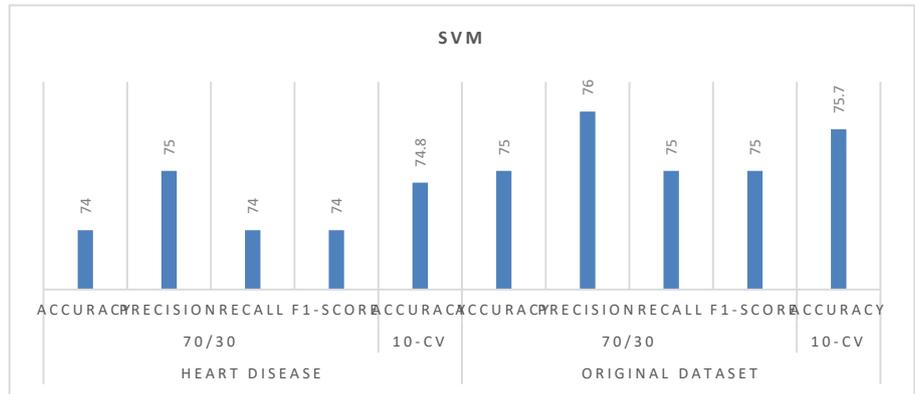

**Figure 17.** SVM Results of the Original Dataset

Figures 18, 19, 20, 21, and 22 present a bar chart comparison of overall algorithms Accuracy, Precision, Recall, and F1-score using a 70/30 approach and Accuracy using a 10-fold CV approach for the original dataset along with four Extracted Datasets.

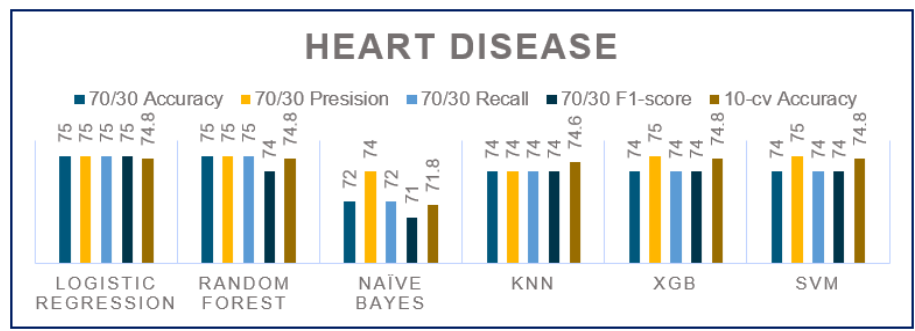

**Figure 18.** Heart Disease Results with CV and 80/20 data split

Elham Musaaed et al.

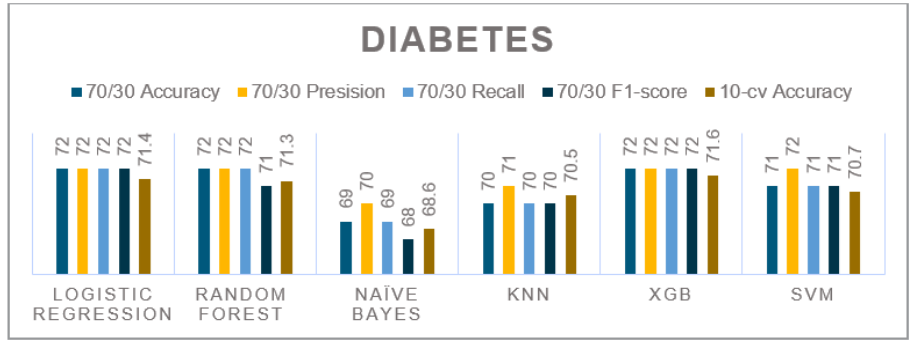

**Figure 19.** Diabetes Results with CV and 80/20 data split

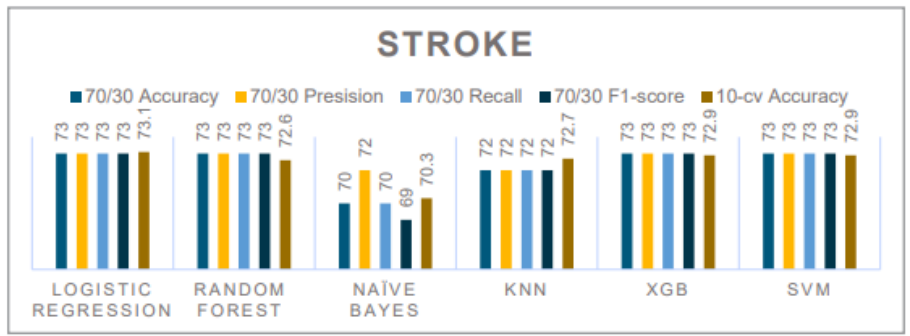

**Figure 20.** Stroke Results with CV and 80/20 data split

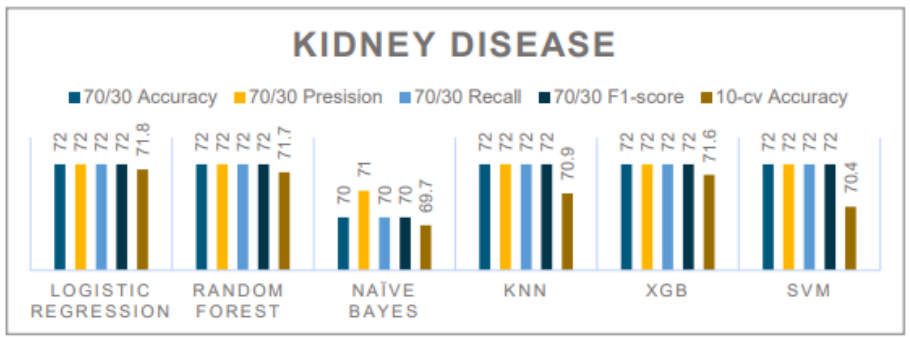

**Figure 21.** Kidney Disease Results with CV and 80/20 data split

Elham Musaaed et al.

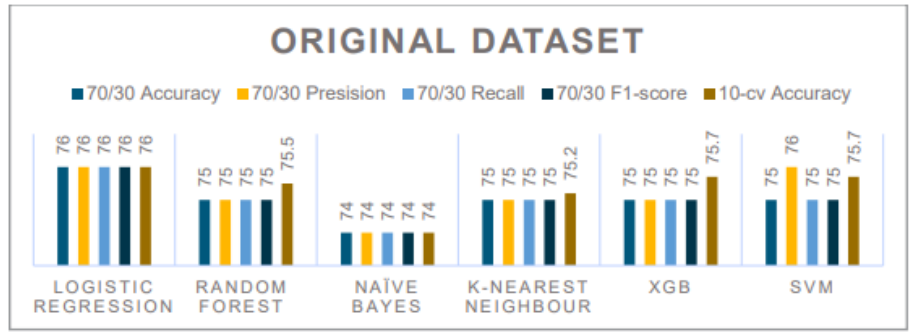

**Figure 22.** Original Dataset Results with CV and 80/20 data split

Although the results were slightly similar between algorithms, the increase was not significant. Therefore, we can conclude that in the Diabetes dataset, LR, XGB, and RF outperformed NB and KNN with 72% accuracy, and in the Stroke dataset, they outperformed NB and KNN by 73%. While the KNN performed slightly better on the kidney dataset to become in line with LR, XGB, and RF, with an accuracy of 72%. In addition, the LR and RF have been found to outperform other algorithms when it comes to the Heart dataset with 75% accuracy. The LR, however, demonstrated superior accuracy in comparison to other Machine Learning algorithms used in the original dataset with 76% accuracy. Furthermore, NB has the lowest performance for all datasets, especially when used to predict diabetics.

To further test the diagnostic ability of the algorithms, ROC-AUC has been applied. The closer the value to 1, the better the performance of the classifier is. In Table 4, we present the ROC-AUC values for the experimented Machine Learning algorithms. The original dataset and the heart dataset had the highest AUC above 0.80. Despite the similarity in the results, we can conclude that NB didn't perform as good as the other ML algorithms. Additionally, LR, RF, and XGB outperformed other ML algorithms in heart, Stroke, and original datasets. In contrast, the LR and RF were higher for the kidney dataset. Furthermore, the LR and XGB were superior in the Diabetes dataset predictions. As with the other measurements, we can draw the same conclusion.

**Table 4** ROC-AUC Results

| Datasets \ ML technique | LR | NB | KNN | RF | XGB | SVM |
|---|---|---|---|---|---|---|
| **Heart** | 0.82 | 0.80 | 0.81 | 0.82 | 0.82 | 0.82 |
| **Diabetes** | 0.79 | 0.75 | 0.77 | 0.78 | 0.79 | 0.78 |
| **Stroke** | 0.79 | 0.78 | 0.78 | 0.79 | 0.79 | 0.79 |
| **Kidney** | 0.79 | 0.78 | 0.78 | 0.79 | 0.78 | 0.78 |
| **Original** | 0.83 | 0.81 | 0.82 | 0.83 | 0.83 | 0.83 |

## 5. Discussion

All algorithms provide the highest prediction of HD when compared with Diabetes, Stroke, and KD when the PRF were used as attributes. Additionally, although there was no significant difference between the algorithms, LR outperformed other ML algorithms with 75% accuracy. As a result of these findings, it appears that these Patient-Related Factors are more closely related to Heart Disease and can be useful in predicting Heart Disease proactively. These findings support the fact that the data was collected for the purpose of identifying the factors that contribute to Heart Disease. However, to improve the accuracy of the prediction of HD, more factors must be considered.

While when Diabetes, Strokes, and KD were included as attributes along with PRF, we found that all algorithms performed better. Based on these findings, Diabetes, Stroke, and KD may contribute to a better prediction of HD. Having a better prediction means that these diseases have an impact on HD. The risk of HD increases with the presence of these diseases, especially Diabetes and Stroke. As indicated by WHO's findings, Diabetes is a leading cause of blindness, KF, heart attacks, Strokes, and lower limb amputations [1]. For adults with Diabetes, the risk of heart attack and Stroke is two- to three-fold higher [2]. This information, combined with the patient's life behaviors, can be used to determine whether the individual is at risk for HD so the appropriate preventive measures are taken to prevent the condition.

On the other hand, the machine learning performance comparison results showed that LR performed better than other algorithms. A linearly separable dataset and a large number of records are essential for LR to perform well. It is also a Predictive Analysis that explains how dependent variables relate to independent variables. NB algorithm is however not performing equal to the other algorithms since it is a simple approach that only works well with small datasets. In addition, NB assumes that all features are independent.

In terms of time consumption, LR and NB consume less time than other algorithms. In RF, the highest score is selected among several independent DT. Having many trees can slow down the algorithm and make it ineffective for making real-time predictions. The XGB algorithm is similar, but it creates a number of trees that are sequentially dependent on the previous trees. SVM is suitable for small and medium-sized datasets due to its kernel-based structure as most implementations store this information as an NxN matrix of distances between the training points in order to avoid having to calculate entries repeatedly. KNN is calculated by comparing the unknown class data with data in the training set and performing a distance measurement. Therefore, KNN will be slower than LR and NB due to its real-time execution.

In light of the overfitting risk associated with some machine learning algorithms and the lack of interpretability, LR performs better when used with large datasets to examine the relationship between the dependent and independent variables. However, comparing LR to other classifiers in terms of complexity, it is a simple algorithm. It is an easy-to-interpret algorithm with fast training time. However, SVM is a more complex algorithm (non-linear). There is a possibility that it will provide better performance, but it may also suffer from overfitting. It is effective with many features and not too many instances, but not when dealing with large data, such as the ones encountered in this study. The performance of RF and XGB in the extracted datasets is comparable to that of logistic regression.. Furthermore, they are not easily interpretable in addition, they require more computing resources when large datasets are considered. If compared with KNN, which is one of the simplest ML algorithms, its complexity lies in the time required to process a single query point O(nd), where n is the number of training examples, and d is the number of features. For each query point, the algorithm must calculate the distance between the query point and every other point within the dataset.

In this study, we only used traditional Machine Learning algorithms such as LR, RF, and KNN, and we did not use DL algorithms. Additionally, we cannot get results from Fully Connected Neural Networks (FCNN) due to the huge datasets and long training times. The dataset was obtained from the CDC to assess the health status of U.S. residents. Nevertheless, this dataset only includes samples related to people in the USA. Lastly, this study examined limited number of PRF, such as age, smoking, sleeping time, general health, and physical activity, as well as Diabetes, Stroke, and KF.

We may extend this work in the future to include DL algorithms such as Convolutional Neural Network (CNN) or FCNNs as well as other ML classifiers. Alternatively, a clustering technique can also be used first, and then apply ML classifiers to see if the accuracy would increase. Furthermore, there is also the possibility of integrating the algorithms with one another or with optimization techniques in a hybrid modeling approach. Also, increasing the number of features such as blood pressure may enhance performance. Lastly, Bahraini hospital data could be used to repeat this experiment.

The main contribution of this comparative study is to examine the influence of patient lifestyle and behaviors on the occurrence of a medical condition by using Machine Learning algorithms in isolation of clinical factors. In most studies, clinical factors have been the primary focus. However, in this study, only Patient-Related Factors were considered, which include age, smoking, sleep time, general health, physical activity, physical health, mental health, alcohol consumption, difficulty walking, gender, race, and body mass index. Several diseases can be attributed to a person's lifestyle. Additionally, Diabetes, Stroke, and KF were examined in conjunction with PRF in order to assess their impact on the presence of HD. Furthermore, to specify the most effective classifier that can be used in this field as well as to develop a web-based application on this classifier that allows users to easily assess their habits and behaviors by submitting their information and receiving a prediction of whether they have a certain medical condition or not.

## 6. Conclusion

In conclusion, all algorithms are effective in predicting HD compared to Diabetes, Stroke, and Kidney Disease. We can therefore conclude that these features are most associated with HD compared to other diseases. This is consistent with the fact that the data was collected for the purpose of studying HD. Moreover, we found that performance did not differ significantly between algorithms, although LR slightly outperformed others, whereas the XBG, RF, and KNN were almost equal regarding the original dataset, while for the extracted dataset the LR, RF, XGB, and KNN had almost similar performance. Furthermore, it can be concluded that LR was always one of the superior classifiers in both the extracted datasets and the original dataset that performed similarly to more complex Machine Learning algorithms, while NB performed the worst. This is due to the fact that LR is a predictive analysis that explains the relationship between a binary dependent variable and a continuous and discrete independent variable. The distribution of classes in feature space is not assumed [23]. Additionally, it provides a probabilistic perspective on class predictions [24].

This study focused on evaluating the performance of the individual ML classifiers in predicting different diseases, once by examining the impact of PRF only on pre-specified diseases and once by examining the impact of including these pre-specified diseases along with PRF as attributes. Using a different approach, this experiment attempts to improve classifiers' performance. As part of this study, we also examined whether the applied classifiers were useful in correctly predicting diseases. Additionally, we developed a Flask-based application on the classifier. This study uses medical datasets, so the classification capability must be excellent since errors could have serious consequences.

In order for medical professionals to make informed decisions regarding the occurrence of diseases, as well as to provide patients with a better understanding of how behavior and lifestyle impact their health, an accurate tool is needed. For the purpose of identifying the most accurate Machine Learning technique for predicting different diseases, comparative studies were conducted. They focused, however, on clinical indicators and medical examinations. We are interested to know how a patient's behavior and habits may contribute to a future medical condition. Furthermore, how Diabetes, Strokes, and KF, among other factors, contribute to HD. In both the extracted datasets and the original dataset, the accuracy of the algorithms ranged from 70% to 76%.

## Acknowledgements

The authors would like to express their gratitude to Dr. Mohammed Ashraf (Doctor in the Ministry of Health) for clarifying the difference between personal indicators and personal habits.

## Conflicts of Interest

All authors have completed the ICMJE uniform disclosure form. The authors have no conflicts of interest to declare.